\documentclass{bmvc2k}

\usepackage{times}
\usepackage{epsfig}
\usepackage{graphicx}
\usepackage{amsmath}
\usepackage{amssymb}
\usepackage{hyperref}
\usepackage{algorithm}
\usepackage[noend]{algpseudocode}

\usepackage{color}
\usepackage{comment}

\newcommand{\beginsupplement}{%
        \setcounter{table}{0}
        \renewcommand{\thetable}{S\arabic{table}}%
        \setcounter{figure}{0}
        \renewcommand{\thefigure}{S\arabic{figure}}%
     }

\DeclareMathOperator*{\argmin}{argmin}

\title{Centroid Based Concept Learning for RGB-D Indoor Scene Classification}

\addauthor{Ali Ayub}{aja5755@psu.edu}{1}
\addauthor{Alan R. Wagner}{alan.r.wagner@psu.edu}{1}
\addinstitution{
 The Pennsylvania State University\\
 State College, PA, USA
}
\runninghead{Ayub et al.}{CBCL for RGB-D Indoor Scene Classification}

\begin{document}

\maketitle
\begin{abstract}
This paper contributes a novel cognitively-inspired method for RGB-D indoor scene classification. High intra-class variance and low inter-class variance make indoor scene classification an extremely challenging task. To cope with this problem, we propose a clustering approach inspired by the concept learning model of the hippocampus and the neocortex, to generate clusters and centroids for different scene categories. Test images depicting different scenes are classified by using their distance to the closest centroids (concepts). Modeling of RGB-D scenes as centroids not only leads to state-of-the-art classification performance on benchmark datasets (SUN RGB-D and NYU Depth V2), but also offers a method for inspecting and interpreting the space of centroids. Inspection of the centroids generated by our approach on RGB-D datasets leads us to propose a method for merging conceptually similar categories, resulting in improved accuracy for all approaches.
\end{abstract}

\section{Introduction}
\label{sec:Intro}
\noindent Classifying images taken from indoor scenes is an important area of research. The development of an accurate indoor scene classifier has the potential to improve indoor localization and decision-making for domestic robots, offer new applications for wearable computer users, and generally result in better vision-based situation awareness thus impacting a wide variety of applications.   
The introduction of deep learning methods, the creation of numerous large-scale datasets, and the development of specialized computing hardware have all contributed to the rapid improvement in image classification performance. One reason for deep learning's success has been the ability to learn multiple layers of generic image features that can then be used on other related computer vision problems. For instance, features from object trained image classifiers have been used to train indoor scene classifiers \cite{Wang_2016_CVPR}. 

\begin{figure}[t]
\centering
\includegraphics[scale=0.3]{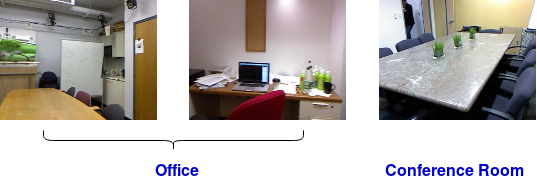}
\caption{\small Indoor scene images have large intra-class variation and small inter-class variation. Sample images are from NYU Depth V2 dataset.}
\label{fig:diff_layouts}
\end{figure}

Yet, indoor scene classification is a challenging problem on its own. Although the presence of certain objects may provide evidence that an image is from a scene category, realistic indoor scenes are often cluttered with numerous objects that are unrelated to the scene's category. Moreover, images taken from an indoor scene often lack category specific information or include information that could be from several different categories. Indoor scene datasets also tend to have an uneven distribution of images across scene categories. Finally, a single scene category (i.e. office in Figure \ref{fig:diff_layouts}) may include a variety of layouts that are composed of different objects and orientations. In general, each image from a scene category can only represent a specific part of the scene containing specific objects. Overall, indoor scenes tend to have high intra-class variation and low inter-class variation (Figure \ref{fig:diff_layouts})  

The development of low-cost RGB-D sensors, such as Microsoft's Kinect, has generated additional interest in indoor scene classification. It has been argued that depth data might offer more robust geometric cues which would be particularly valuable for indoor scene classification \cite{Zhu_2016_CVPR}. Recent research has successfully used this depth information to learn an additional set of discriminating features which, when used with the RGB features, improves indoor scene classification accuracy \cite{Song19}. State-of-the-art approaches for indoor scene classification thus focus on developing methods that correlate features across the RGB and depth modalities and learn the relationships of features across modalities, while classification is performed using either a softmax classifier or SVM \cite{Li18,Yuan19}.

% Ali, this section essentially repeats the same thing that is in other parts of the paper.

In contrast to these prior approaches, this paper presents a cognitively-inspired framework called CBCL (Centroid-Based Concept Learning) for RGB-D indoor scene classification. %CBCL is related to the concept learning model of the hippocampus and the neocortex \cite{Mack17,moscovitch16,renoult15}. 
%CBCL uses the Places365-CNN ~\cite{zhou2017} to extract a feature map for the RGB modality and a fine-tuned Places365-CNN to extract a feature map for the depth modality. Then for each scene category, CBCL finds a set of centroids using a cognitively-inspired clustering (termed \textit{Agg-Var} clustering in this paper) approach on the RGB-D feature maps, such that each centroid captures different layouts of the scene category. 
CBCL calculates a set of centroids using a cognitively-inspired clustering (termed \textit{Agg-Var} clustering) method on RGB-D feature maps, resulting in centroids that capture different layouts of the same scene category. Classification of an unlabeled RGB-D image is achieved by finding the distance to a set of closest centroids. One major advantage of CBCL is that the resulting space of clusters can further be used to analyze the data and identify and merge conceptually similar categories. Furthermore, it can applied to other classification tasks, such as incremental learning of objects from object-centric RGB images \cite{Ayub_2020_CVPR_Workshops,Ayub_IROS_20,Ayub_ICML_20}. The complete code for this paper is available at: \url{https://github.com/aliayub7/CBCL_RGBD}

The main contributions of this paper are: 1) A cognitively-inspired method for RGB-D scene classification approach that achieves state-of-the-art classification accuracy on benchmark datasets. 2) A demonstration that internal cluster evaluations can be used directly with CBCL to update the cluster structure by combining conceputally similar categories further improving classification accuracy for all approaches. % use the output of CBCL to analyze the benchmark datasets and suggest updated versions of the datasets for RGB-D scene classification.
3) The time required to train a model using CBCL is much lower than training an additional deep network.
%The remainder of the paper begins by presenting the related work, describing the algorithm, presenting experimental evaluations of the method, and conclusions. 

%The main contributions of this paper are: 
%\begin{enumerate}
%    \item A cognitively-inspired RGB-D indoor scene classification approach that deals with high intra-class variance effectively and achieves state-of-the-art results.
%    \item An approach that produces an \textit{interpretable} model. Lack of model interpretability has been one criticism of traditional deep learning \cite{Lipton17}.
%    \item The time required to learn a model using CBCL is much lower than training an additional deep network.
%\end{enumerate}
%1) A cognitively-inspired RGB-D indoor scene classification approach that deals with high intra-class variance effectively and achieves state-of-the-art results. 2)An approach that produces an \textit{interpretable} model. Lack of model interpretability has been one criticism of traditional deep learning \cite{Lipton17}. 3) The time required to learn a model using CBCL is much lower than training an additional deep network. 

\section{Related Work}
\label{sec:Related}
\noindent 
%Approaches for indoor scene classification have been influenced by different research directions. This section first 
%The related work reviews methods for scene classification, RGB-D scene classification, and clustering-based classification.  

\subsection{Scene Classification} 
Techniques for scene classification have rapidly improved. Early work relied on handcrafted features and low-level spatial information \cite{Szummer98}. Recent approaches tend to favor using Convolutional Neural Networks (CNN) to extract features from indoor scenes \cite{Zhou14}. Zhou et al. \cite{zhou2017} notes that the features extracted from an ImageNet \cite{Russakovsky15} trained CNN result in poor performance on indoor scene classification. They therefore created the Places365 dataset which includes over 10 million labeled images of different scenes. Another approach has been to pool local image features using Fisher Vectors (FV) or Vector of Locally Aggregated Descriptors (VLAD) \cite{Wang17}. The performance of pooled image features, however, is susceptible to the noise inherent in the image patches. Doshi et al. \cite{Doshi15} demonstrate an early system that uses features generated by a CNN pretrained on ImageNet encoded as Fisher Vectors to classify scenes from streaming, first-person video. Unfortunately, this work does not evaluate their method on the standard datasets or compare its performance to other methods. 

\subsection{RGB-D Scene Classification} 
\label{sec:rgbd_related_work}
The availability of low-cost RGB-D sensors has encouraged the development of methods seeking to improve indoor scene classification by using features from both the RGB and depth modalities. In an early approach, \cite{Gupta15} proposed to extract local features by quantizing segmentation outputs and detecting contours on depth images. More recent approaches use CNNs because of their performance on object classification tasks \cite{Krizhevsky12}. Current state-of-the-art approaches focus on developing better methods to represent and correlate the RGB and depth features. %Cheng et al. \cite{Cheng_2017_CVPR} use modality-specific features learned separately from RGB and depth images and then fuse the results at the score level. 
Depth features have also been learned independently and different fusion strategies have been explored in an attempt to maximize performance \cite{Song19}. Li et al. \cite{Li18} presents a classification pipeline that learns and uses a fusion network. Yuan et al. \cite{Yuan19} achieves improved classification accuracy on the SUN RGB-D dataset using a cross-modal graph convolutional network to capture and use the RGB and depth relationships. These approaches have improved RGB-D scene classification by mainly focusing on learning better scene representations for the RGB and depth modalities. We present an entirely different classification approach which clusters the deep CNN features of the RGB-D images using a cognitively inspired approach, that leads to further benefits than simply improving accuracy.        

\subsection{Clustering-based Classification}

Combinations of clustering and deep neural networks have been developed in the past \cite{DBLP:journals/corr/abs-1801-07648}. Yang et al. \cite{yang2016joint} develops an unsupervised method for joint CNN and agglomerative clustering representation learning in which CNN features are used as an input to the agglomerative clustering procedure. Their method demonstrates near state-of-the-art performance on several object detection datasets. With respect to supervised machine learning, clustering has mostly been used for text classification tasks in recent years \cite{Liu17}. Although, k-nearest neighbors has been applied for object-centric image classification tasks \cite{Meyer18}, to the best of our knowledge, the proposed \textit{Agg-Var} clustering approach with CNN features has not been applied to the problem of RGB-D indoor scene classification.  
\section{Methodology}
\label{sec:Method}

\begin{figure}[t]
\centering
\includegraphics[scale=0.22]{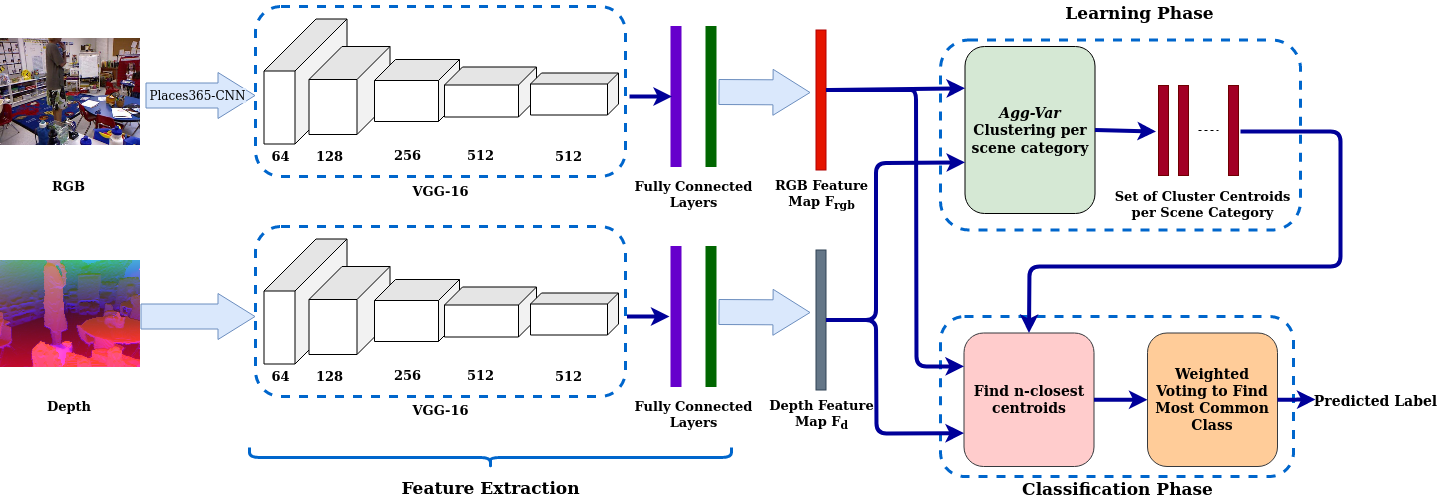}
%\smallskip
\caption{\small The framework for our proposed approach generates the centroids for each scene category using \textit{Agg-Var} clustering and uses the centroids for classifying unlabeled RGB-D images.}
\label{fig:framework}
\end{figure}

\noindent CBCL is cognitively inspired from concept learning models of the hippocampus and the neocortex \cite{Mack17,moscovitch16,renoult15}. According to the concept learning literature, whenever a new episode is encountered by the hippocampus it first extracts feature information about the episode. The difference from the feature map of the episode to previously learned concepts, termed \textit{memory-based prediction error}, is then calculated. When this difference is large, the hippocampus performs \textit{pattern separation} by creating a new distinct concept for the episode. For smaller differences, \textit{memory integration} is performed by updating an existing concept to incorporate the new episode. CBCL learns concepts pertaining to different layouts of scene categories in the form of centroids. The subsections below describe the algorithm in detail. %explain CBCL's centroid generation process and classification technique in detail.

%Figure \ref{fig:framework} graphically depicts our approach. The first step extracts features for the RGB modality using VGG-16 CNN \cite{Karen14} pre-trained on the Places365 dataset (Places365-CNN) and for the depth modality using a fine-tuned Places365-CNN on the depth data. Next the \textit{Agg-Var} clustering approach is applied to the training dataset for each category of scene resulting in a set of centroids for each scene category. Finally, the centroids are used to classify unlabeled scene images. 

\subsection{\textit{Agg-Var} Clustering}
Our approach (Figure \ref{fig:framework}) begins with the generation of separate features maps for the two modalities. A VGG16 CNN~\cite{Karen14} pre-trained on the Places365 dataset~\cite{zhou2017} is used to generate features map for the RGB modality and a VGG16-CNN trained from scratch on depth data is used to generate feature maps for the depth (HHA encoded) modality. Feature maps are extracted from the second fully connected layer of VGG16. Using the notation of \cite{Yuan19}, we denote input RGB data as $x_{rgb}$ and input depth data as $x_d$ and the feature maps generated by the second fully-connected layer of the VGG network as $F_{rgb}$ and $F_d$.

In the learning phase, \textit{Agg-Var} clustering is applied to each feature map in a scene category. %After the feature extraction step, the concept learning process in the hippocampus calculates a term called the \textit{memory-based prediction error}. This value represents the difference from the incoming episode to all of the previously experienced concepts. We replicate this step by finding the overall distance from an incoming RGB-D image to  each centroid for a scene category. 
Similar to the first step in the hippocampal concept learning model, our approach finds the overall distance from an incoming RGB-D image to each centroid for a scene category. Initially there are no centroids in a scene category. Hence, this step begins by creating two centroids, one for each modality, from the first image in a category. Next, for each image in all categories, feature maps $F_{rgb}$ and $F_d$ are generated and compared using the Euclidean distance to all the centroid pairs in the category. A weighted average of the Euclidean distances for each modality is used to calculate the overall RGB-D distance, $dist_{RD}(c_{rd_i}^j,F_{rd})$, between the $i$th centroid pair $c_{rd_i}^j$ (for both modalities) in the scene category $j$ and the feature map pair $F_{rd}$:
\begin{align}
dist_{RD}(c_{rd_i}^j,F_{rd})= \frac{1}{2}\biggl(w_{rgb}\times dist(c_{rgb_i}^{j},F_{rgb}) %\notag\\
+ w_D\times dist(c_{d_i}^j,F_d)\biggr)
\end{align}

%\noindent where,
%\[dist(F,F^{'}) =\sqrt{\sum_{l=1}^{z} (F_{l}-F^{'}_{l})^2}\] 

\noindent where $dist(c_{rgb_i}^{j},F_{rgb})$ and $dist(c_{d_i}^j,F_d)$ are the Euclidean distances between RGB feature map $F_{rgb}$ and $i$th RGB centroid $c_{rgb_i}^j$ and depth feature map $F_d$ and $i$th depth centroid $c_{d_i}^j$ for the scene category $j$, respectively. %is the euclidean distance between two feature maps $F$ and $F^{'}$ of dimension $z$ and $c_{rgb_j}^i$ and $c_{d_j}^i$ are the $i$th RGB and depth centroids for the scene category $W_j$. 
The weights $w_{rgb}$ and $w_D$ are hyper-parameters we term \textit{fusion weights} with values ranging from 0 to 1, for the RGB and depth distances. For each image in all categories, the distance $dist_{RD}(c_{rd_i}^j,F_{rd})$, is computed for all current centroid pairs in a category to the image feature map pair, $F_{rd}$.

If the distance $dist_{RD}(c_{rd_i}^j,F_{rd})$ to the closest centroid pair is below a pre-defined distance threshold $D$, the centroid for each modality is updated by calculating a weighted mean of the centroids and the feature maps of the new image:

\begin{equation}
C_{new} = \frac{w_C*C_{Old} + F}{w_C+1}
\end{equation}

\noindent where, $C_{new}$ is the updated centroid, $C_{old}$ is the centroid before the update, $w_C$ is the number of data points (images) already represented by the centroid and $F$ is the feature map of the new image. Equation (2) is used to calculate the updated centroids for both the RGB and depth modalities. This step of our approach is meant to capture the \textit{memory integration} portion of the concept learning process of the hippocampus. If, on the other hand, the overall distance between the new image and the nearest centroid is higher than the distance threshold $D$, a new centroid pair (for both modalities) is created for that category and equated to the feature map pair for the image. This step is inspired by the \textit{pattern separation} procedure in the hippocampal concept learning process. %, in which a new distinct concept is created for an incoming episode.
%This step of our process is meant to capture the \textit{memory integration} step in the EpCon model. Memory integration occurs when the \textit{memory-based prediction error} of an episode to a previous concept is small. When this is the case the incoming episode is integrated into an existing concept. If, on the other hand, the \textit{memory-based prediction error} of an episode to a previous concept is large, according to the EpCon model, \textit{pattern separation} occurs resulting in the creation of a new distinct concept based on the incoming episode. Our approach captures this aspect of the model as: when the overall distance between the new image and the nearest centroid is higher than the distance threshold $D$, a new centroid pair (for both modalities) is created for that category and equated to the feature map pair for the image. 
This process is repeated for each image in the training data for each category. The result of this process is a collection containing a set of centroid pairs $C_{RD}=\{C_{RD}^1,C_{RD}^2,...,C_{RD}^m\}$ for $m$ indoor scene categories.

An argument can be made against CBCL that with the increase in the number of training images, the number of centroids will increase resulting in memory and computation issues. However, an increase in the number of training images does not necessarily mean that the number of centroids will increase because new training images with similar layouts to previous images will be merged to existing centroids. If the system does run out of memory, we can reduce the total number of centroids by applying Agg-Var clustering on the existing centroids with a larger distance threshold, thus combining similar centroids into a single centroid, decreasing the memory used \cite{Ayub_2020_CVPR_Workshops}.  

\subsection{Classification of Unlabeled RGB-D Images}
%The output from \textit{Agg-Var} clustering, a collection $C_{RD}$ containing a set of centroid pairs for each category of indoor scenes, is used to classify unlabeled images. 
To classify an RGB-D image, features maps $F_{rgb}$ and $F_d$ are first extracted from the CNN. Next, equation (1) is used to calculate the distance between the unlabeled image feature maps and the centroid pairs in $C_{RD}^j$ for each scene category $j$. Based on the calculated distances, $n$ closest centroid pairs to the unlabeled image are selected. Each of the $n$ closest centroid pairs contributes to the prediction of category $j$ according to the conditional summation,
%The contribution of each of the $n$ closest centroid pairs to the determination of a scene category $j$ is a conditional summation:

\begin{equation}
    Pred(j) = \sum_{i=1}^{n} \frac{1}{dist_{RD_i}}[y_i=j]
\end{equation}

\noindent where $Pred(j)$ is the prediction weight for the scene category $j$, $y_i$ is category label of $i$th closest centroid pair and $dist_{RD_i}$ is the distance (calculated using Eq. 1) between the $i$th closest centroid pair and the feature maps for both modalities of the test image. The prediction weights for all the categories are initialized to zero. Then, for the $n$ closest centroid pairs the prediction weights are updated for the categories that each of the $n$ centroid pairs belong to. The prediction weight for each class is further multiplied by the inverse of the total number of images in the training set of the class to deal with class imbalance. The test image is classified based on the category with the highest prediction weight. %From equation (3), the prediction weight of a category is directly proportional to the number of centroid pairs, among the $n$ centroid pairs, that belong to the category, and inversely proportional to the distance of those centroid pairs from the test image's feature maps. Hence, the closest centroid pair contributes the most to the prediction of the category of the test image. The hyper-parameter, $n\geq 1$, was chosen empirically. Intuitively, because a scene category typically has multiple layouts, a test image can have different patches of pixels that match to different scene layouts. Hence, more than one closest centroid pairs (representing different layouts of a scene category) are considered when predicting the unlabeled image's category.

\section{Experiments}
\label{sec:Experiments}
\noindent The proposed approach was evaluated on two standard RGB-D scene classification datasets: SUN RGB-D and NYU Depth V2. The datasets are first briefly described and then the performance of our approach is compared to the state-of-the-art methods for indoor scene classification. Finally, we show how approach can be used to merge similar categories in RGB-D datasets resulting in improved classification accuracy.

\subsection{Datasets}
The SUN RGB-D dataset is the largest publicly available dataset for RGB-D indoor scene classification. It includes 10,335 RGB and depth image pairs captured from a variety of different camera and depth sensors. To be consistent with our predecessors' experimental setup, we use 19 categories, all of which have more than 80 images \cite{Song_2015_CVPR}. We also keep the standard splits, with 4,845 images for training and 4,659 images for testing.

The NYU Depth V2 dataset includes 1,449 RGB and depth image pairs consisting of 27 different categories. Because many categories have few training examples, \cite{Silberman12} reorganized the 27 categories into 10 categories with 9 usual scene types and one "others" category. To again be consistent with our predecessor's experimental setup, we follow the same category settings and the data split settings as in \cite{Silberman12} with 795 training images and 654 test images.

\subsection{Implementation Details}
To generate the feature maps for the images, VGG16 was implemented with the Pytorch deep learning framework \cite{torch19} and initialized with pre-trained Places365 weights for RGB modality. A TITAN RTX GPU was used for feature extraction and training and a Ryzen Threadripper 1920x CPU was used to create the centroids and classify test images. The input images were resized to $256 \times 256$ and randomly cropped to $224\times 224$ as the input to the network. For the depth modality of the SUN RGB-D dataset, we trained a VGG16 from scratch for 200 epochs and for the NYU Depth V2 we fine-tuned the VGG16 network trained on the SUN RGB-D depth dataset for another 200 epochs. We also tested by finetuning the Places365 CNN on depth data but it produced lower accuracy than the VGG16 trained from scratch. In accordance with the previous approaches \cite{Yuan19,Du_2019_CVPR}, we report top-1 average accuracy over all scene categories. The hyperparameters for CBCL (distance threshold $D$, \textit{fusion weight} $w_D$ and number of closest centroids for classification $n$) were tuned using cross-validation on the training set. The hyperparameter values for CBCL and training the neural networks are provided in the supplementary file.

\subsection{Results on the SUN RGB-D Dataset}
\label{sec:SUN_results}
Table \ref{tab:SUNRGBD} compares our approach (CBCL) to six state-of-the-art (SOTA) methods on the SUN RGB-D dataset. Most of these methods rely on fine tuning of the Places205 CNN (AlexNet pretrained on Places205 dataset). All of the SOTA methods are described in section \ref{sec:rgbd_related_work}. We use Places365 features for our approach because it is a bigger and newer version of Places205. We therefore hoped that Places365 features would be better than Places205, however they produce similar accuracy as Places205 features (see section \ref{sec:ablation_study} below).% and are even inferior to Places-205 features on some datasets \cite{zhou2017}.

\begin{table}
\centering
\small
\begin{tabular}{ |p{3.9cm}|p{1.0cm}|p{1.0cm}|p{1.0cm}| }
     \hline
    \textbf{Methods} & \textbf{RGB} & \textbf{Depth} & \textbf{Fusion} \\
    \hline
    VGG Baseline & 42.5 & 35.6 & 49.8\\
    \hline
    AlexNet Baseline & 42.6 & 38.4 & 48.3\\
    \hline
    ResNet18 Baseline & 47.4 & 44.8 & 50.8\\
    \hline
    Agglomerative RGB-D & 45.6 & 34.0 & 53.8\\
    \hline
    K-means RGB-D & 41.8 & 31.6 & 47.9\\
    \hline
     \hline
    %Song et al. \cite{Song_2015_CVPR} & -- & -- & 39.0\\
     %\hline
    %Liao et al.\cite{Liao16} &  36.1 & -- & 41.3\\
    %\hline
    %Zhu et al. \cite{Zhu_2016_CVPR} &  40.4 & 36.5 & 41.5\\
    %\hline
    Wang et al. \cite{Wang_2016_CVPR} & 40.4 & 36.5 & 48.1\\
    %\hline
    %Song et al. \cite{Song17} & - & 40.1 & 52.3\\
    \hline
    Du et al. \cite{Du18} & 42.6 & 43.3 & 53.3\\
    \hline
    Li et al. \cite{Li18} & 46.3 & 39.2 & 54.6\\
    \hline
    Song et al. \cite{Song19} & 44.6 & 42.7 & 53.8\\
    \hline
    Yuan et al. \cite{Yuan19} & 45.7 & -- & 55.1\\
    \hline
    Du et al. \cite{Du_2019_CVPR} Aug. & \textbf{50.6} & \textbf{47.9} & 56.7\\
    \hline
    \textbf{CBCL} & 48.8 & 37.3 & \textbf{59.5}\\
    %\hline
    %\textbf{CBCL} Resnet\_Depth & 48.8 & 38.5 & \textbf{61.2}\\
 \hline
 \end{tabular}
 \bigskip
 \caption{Comparison with baselines and state-of-the-art methods on \textbf{SUN RGB-D} test set. Performance depicted as classification accuracy(\%). Aug. denotes use of augmented data.}
 \label{tab:SUNRGBD}
 \end{table}
% Actual 75, 16, 0.79, 66.2%
% baseline 55.2%

CBCL outperforms these prior state-of-the-art methods on RGB-D evaluation by \textbf{2.8\%} achieving an accuracy of \textbf{59.5\%} (see Table \ref{tab:SUNRGBD}). For a fair comparison with \cite{Du_2019_CVPR}, we tested CBCL with ResNet18 backbone (CBCL ResNet18 in Table \ref{tab:SUNRGBD}) which produced \textbf{61.2\%} accuracy, \textbf{4.5\%} higher than \cite{Du_2019_CVPR}. Note that \cite{Du_2019_CVPR} still has an advantage of using augmented data. CBCL also beats other methods on RGB only evaluations except \cite{Du_2019_CVPR} which has the advantage of using augmented data and ResNet18 backbone. Our accuracy on the depth modality is low because we are simply using a VGG16 trained from scratch on depth data. We have intentionally made little effort to optimize depth features, choosing instead to focus on the value added by our classification approach.

\begin{comment}
\begin{table}
\centering
\small
\begin{tabular}{ |p{6.1cm}|p{1.15cm}| }
     \hline
    \textbf{Methods} & \textbf{Accuracy (\%)}\\
     \hline
     RGB VGG & 42.47\\
     \hline
     CBCL RGB ($n=24$, $D=100$) & \textbf{48.82}\\
     \hline
     Depth(HHA) VGG & 35.58\\
     \hline
     CBCL Depth(HHA) ($n=12$, $D=95$) & \textbf{36.22}\\
     \hline
     RGB-D(HHA) VGG & 49.84\\
     \hline
     CBCL RGB-D(HHA) ($n=13$, $D=85$, $w_D=0.70$) & \textbf{57.84}\\
     \hline
     Agglomerative RGB-D(HHA) ($n=1$, $D=30.4$, $w_D=0.70$) & \textbf{56.05}\\
\hline
\end{tabular}
\caption{Ablation study on \textbf{SUNRGB-D} dataset}
\label{tab:SUN_ablation}
\end{table}
\end{comment}

The time required to train a VGG16 on the depth modality is approximately 2 hours. The time required to generate the feature maps for all the training images is $68.09s$ and the time required to generate the centroids is only $1.03s$. Hence our method is significantly faster than training an additional deep network. 

\subsubsection{Ablation Study}
\label{sec:ablation_study}
\noindent Table \ref{tab:SUNRGBD} also presents the results for an ablation study of our method. We created three different baselines: the VGG baseline is generated by using a fine-tuned VGG16 network pre-trained on the Places365 dataset. The AlexNet baseline uses a fine-tuned AlexNet network pre-trained on the Places205 dataset, while the ResNet18 baseline uses a fine-tuned ResNet18 network pre-trained on the Places205 dataset. All three baselines use streams for each modality concatenated at the last fully connected layer. Next, two hybrids of our approach were created by replacing \textit{Agg-Var} clustering with two different clustering approaches: traditional agglomerative clustering \cite{gowda1978agglomerative,gdalyahu2001self} and K-means clustering. 

The VGG and AlexNet baselines perform similarly on the RGB, depth and RGB-D evaluations, although the ResNet18 baseline performs the best among the three. These results show that choice of VGG16 and Places365 features does not offer an unfair advantage to our approach over other architectures or features. CBCL improves performance on RGB and RGB-D evaluations over the three baselines by significant margins. For the depth modality, however, it beats the VGG baseline but is inferior to ResNet18 and AlexNet baselines because it uses VGG16 backbone. Both the agglomerative and K-means clustering hybrids of CBCL are inferior to CBCL with \textit{Agg-Var} clustering on the RGB, depth, and fusion evaluations, demonstrating the contribution of \textit{Agg-Var} clustering towards CBCL's performance. The reason being, K-means finds a fixed number of centroids for each category using hyperparameter k, which prevents some layouts from being optimally represented, especially when intra-class variance is high. Agg-Var uses a similarity threshold to dynamically combine centroids, thus generating a different number of centroids for different categories using a single hyperparameter. Agglomerative clustering also uses a similarity threshold; however it uses a pre-defined distance matrix for clustering instead of calculating distances directly on centroids and merging them continually like Agg-Var. Hence, we believe that agglomerative clustering does not capture the dynamic changes in centroids caused by merging data points, because, with additional data, the updated distances in the distance matrix for the clusters start to drift from actual distances between centroids.

\begin{figure}[t]
\centering
\includegraphics[scale=0.28]{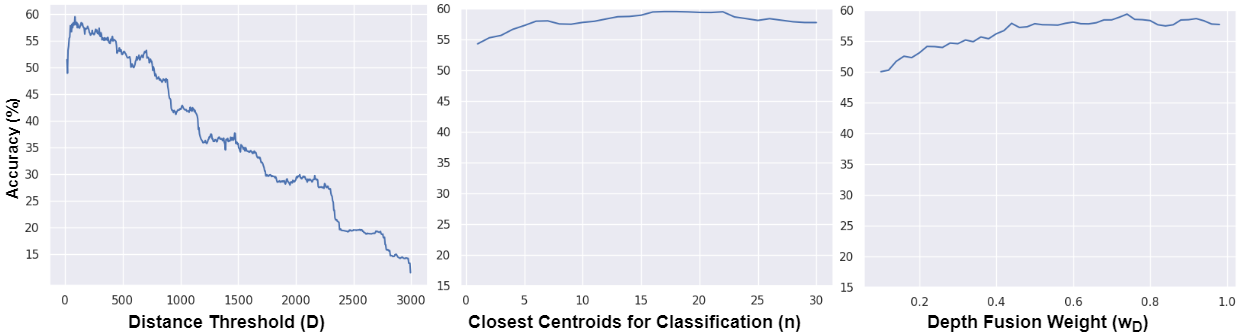}
\caption{\small The effects of varying distance threshold ($D$), the ($n$) closest centroids used for classification and \textit{fusion weight} ($w_D$) on the classification accuracy on the test set of \textbf{SUN RGB-D} dataset. While changing one of the three parameters, the other parameters' values were set to be: $D=85$, $n=17$, $w_D=0.73$ for the best results}
\label{fig:parameter_analysis}
\end{figure}

\subsubsection{Hyperparameter Analysis}
\label{sec:hyperparameter_analysis}
Figure \ref{fig:parameter_analysis} depicts impact of different hyperparameter values for $D$, $w_D$ and $n$ on classification accuracy. For the distance threshold $D$, smaller values result in each centroid being composed of one or a few images which leads to overfitting. Too large threshold, on the other hand, results in too few centroids to represent all the scene layouts in each category. Still, for a large range of $D$ values (80-200) the classification accuracy stays within a 1\% range. Variations of the hyperparameter $n$ (number of closest centroids for prediction) over the range 5-30 impacts accuracy by less than 3\%. Similarly variations of fusion weight from 0.5-0.95 also impact accuracy by less than 3\%. The figures demonstrate that the accuracy of the CBCL algorithm is relatively insensitive to the values of these hyperparameters over a large range of possible values.

\subsection{Results on NYU Depth V2 Dataset}
Table \ref{tab:NYU} compares our approach on the NYU Depth V2 dataset to six state-of-the-art methods on both modalities and their fusion. All of the methods being compared to were introduced in section \ref{sec:SUN_results}. Our method outperforms all other methods on the RGB and RGB-D evaluations. As with the SUN RGB-D dataset, our method is not optimized for the depth modality. It should be noted that our method outperforms \cite{Du_2019_CVPR} on both RGB and RGB-D modalities even though they use SUN RGB-D features and augmented data. For a fair comparison, we compare our method with \cite{Du_2019_CVPR} using Places features (instead of SUNRGB-D features). In this case, CBCL outperforms \cite{Du_2019_CVPR} with a much greater margin (\textbf{5.4\%}). An ablation study on NYU Depth-V2 dataset is provided in the supplementary file, and is consistent with the ablation study performed on the SUN RGB-D dataset.

\begin{table}[t]
\centering
\small
\begin{tabular}{ |p{4.2cm}|p{0.8cm}|p{0.8cm}|p{0.8cm}| }
     \hline
    \textbf{Methods} & \textbf{RGB} & \textbf{Depth} & \textbf{Fusion} \\
     \hline
    Wang et al. \cite{Wang_2016_CVPR} & 53.5 & 51.5 & 63.9\\
    \hline
    Song et al. \cite{Song17} & - & - & 66.7\\
    \hline
    Li et al. \cite{Li18} & 61.1 & 54.8 & 65.4\\
    \hline
    Du et al. \cite{Du18} & 53.7 & \textbf{59.0} & 67.5\\
    \hline
    Yuan et al. \cite{Yuan19} & 55.4 & -- & 67.4\\
    \hline
    Song et al. \cite{Song19} & 53.4 & 56.4 & 67.5\\
    \hline
    Du et al. \cite{Du_2019_CVPR} Places & 60.2 & 55.2 & 65.5\\
    %\hline
    %Du et al. \cite{Du_2019_CVPR} SUNRGB-D & 63.8 & 56.7 & 66.5\\
    \hline
    Du et al. \cite{Du_2019_CVPR} SUNRGB-D Aug & 64.8 & 57.7 & 69.2\\
    \hline
    \textbf{CBCL} & \textbf{66.4} & 49.5 & \textbf{70.9}\\%\textbf{69.7}\\
 \hline
 \end{tabular}
 \caption{Accuracy (\%) comparison against state-of-the-art methods on the \textbf{NYU Depth V2} test set. Aug means usage of augmented data, Places means fine-tuning on Places features, SUNRGB-D means fine-tuning on SUN RGB-D features.}
 \label{tab:NYU}
 \end{table}
 
\subsection{Model Inspection and Category Merging}
\begin{figure}[t]
\centering
\includegraphics[scale=0.285]{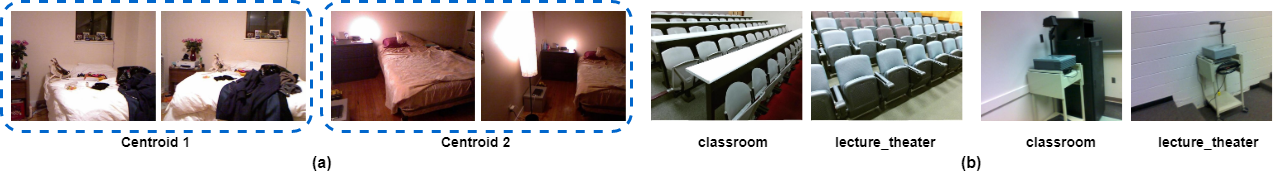}
\caption{\small (a) Images from the category \textit{bedroom} with different layouts are represented by different centroids after \textit{Agg-Var} clustering. (b) Images from the scene categories \textit{classroom} and \textit{lecture\_theatre} are represented by different centroids even though they are very similar. The distance between the corresponding centroids of the images is small, hence the classifier misclassifies these images. Sample images are from SUN RGB-D dataset.}
\label{fig:diff_cetroids}
\end{figure}

\noindent  %We find that images with the same layout are in the same cluster while images with different layouts are in different clusters. %This demonstrates an additional advantage of our approach in model, \textit{interpretability}.
Consistent with the confusion matrices (see supplementary file),  %resulting from application of our method. %Scene categories that are misclassified by CBCL, the distance between the centroids of those classes is small because of the similarity in images that compose the scene categoreis (Figure \ref{fig:diff_cetroids} (b)).
the distance between the centroids of the scene categories that were misclassified is small. Because the output of our approach is a set of centroids, the clusters within a scene category can be examined by looking at the images that compose these centroids (Figure \ref{fig:diff_cetroids} (a)). %representing different layouts of the confused categories is small. 
There is considerable similarity between the layouts that are represented by the centroids of the confused categories (Figure \ref{fig:diff_cetroids} (b)) because the dataset includes similar looking scene layouts labeled as different scene categories (see the supplementary file for additional examples). %Looking at these examples, it is difficult to imagine a visual difference that might be used by a classifier to correctly categorize these images.  
One powerful advantage of our approach is the possibility of formally measuring the internal consistency of the clusters and using this information to reorganize cluster centroids.   

%Some of the categories are so semantically similar that it might require relabeling of such categories by human assistance, or combining them into a single superclass. Perhaps using CBCL's output and nearby centroids to produce groundtruth generating questions for human evaluators. 

The silhouette index ($s$) is one of the methods used to measure internal cluster consistency \cite{Rousseeuw87}. The silhouette value measures how similar an image is to its own cluster, with values below zero indicating that an image is closer to another cluster's centroid than its own. Using the centroids/clusters generated by CBCL, we calculated the silhouette indices of all the training images for the SUN RGB-D dataset. The analysis revealed that some categories in the dataset have a significant percentage of training images ($\geq$25\%) with silhouette value $s\leq 0$. We then looked at the category that was a better suite for the image. The following pairs emerged: (classroom, lecture\_theatre), (conference\_room, study\_space), (living\_room, rest\_space), (home\_office, office) (see supplementary file for a full analysis and algorithm used). This process was conducted recursively. This suggests that there are some categories that are conceptually very similar to each other but have been labeled differently. We therefore combined each of these four category pairs into single categories, reducing the total number of classes for the SUN RGB-D dataset from 19 to 15 and then reevaluated CBCL on this new dataset. The new performance was \textbf{66.2\%}, $6.7\%$ higher than on the original dataset. We also tested the VGG baseline on the updated dataset as well and its accuracy increased to 55.2\%, a 5.4\% increase but still \textbf{$11\%$} below CBCL. One could argue that the accuracy is higher simply because there are fewer categories. We therefore randomly combined 4 category pairs from the dataset and once again reevaluated CBCL and the VGG baseline. The resulting accuracies were similar to the accuracies on the original dataset (0.4\% and 0.3\% increase, respectively), indicating that the increase in accuracy is not solely because of the smaller number of classes. When the same process was used on the NYU V2 dataset, CBCL achieved \textbf{78.95\%} mean class accuracy which is 8.04\% higher than on the original dataset.  

Our process of inspecting the model and merging categories is self-supervised, giving our system the ability to reevaluate the labeling of the data. Poor cluster structure may signal that class categories should be combined. Although doing so may prevent fair comparison to previous methods, category merging may allow for a more realistic conceptualization of the data, especially when the original labeling of the dataset is significantly flawed or inaccurate. The resulting dataset may have greater conceptually consistent labeling which can be used by the future methods as a benchmark. Importantly, our method only merges categories that are visually similar. It assumes that visual similarity predicts conceptual similarity, and this may not be the case. Human assistance can be used in such cases to confirm the updated labels. For applications such as robotics, merging and perhaps separating clusters may offer a dynamic mechanism for capturing the structure underlying the data and increasing classification accuracy, even in the presence of errors.

\section{Conclusion}
\label{sec:Conclusion}
\noindent This paper has introduced a new approach (CBCL) for RGB-D indoor scene classification based on the concept learning model of the hippocampus and the neocortex. The proposed approach not only results in state-of-the-art performance on benchmark datasets but also affords novel methods for interpreting the model and evaluating when to merge categories leading to better overall classification performance.    
%Our method uses the Places365-CNN features and then applies \textit{Agg-Var} clustering on the training data of each scene category to generate unique centroids (concepts) representing different scene layouts for each scene category. To classify unlabeled images, the $n$ closest centroids are used in a weighted voting procedure that selects the best category for the image. Our approach produces state-of-the-art performance on the SUN RGB-D and NYU Depth V2 datasets. %Furthermore, our approach requires less training time than other methods and represents the scene layouts as cluster centroids improving model \textit{interpretability}. 
Our approach can also be applied for other computer vision problems, such as incremental learning. In contrast, most methods are specifically tailored to a particular problem and are not applicable across tasks.  

Our work offers new avenues of research for indoor scene classification. Merging clusters that meet well-defined criteria may offer a means for improved concept learning. Not only could this approach rectify problems with incorrectly labeled data, it might allow for concept generalization, perhaps allowing classifiers to one day generalize across concepts. Our future work will attempt to further develop this aspect of the process.  

%methods that allow a classifier to learn the core features that fit a location to its labeled category. \textcolor{blue}{I am not sure what this line is trying to say? Neural networks already learn features that fit a location to its labeled category.}   

%As the data shows, the success of our approach is hindered by our use of sub-optimal depth features. Our future work will focus on generating better depth features which we expect to dramatically increase classification accuracy.  %Towards this goal we intend to publicly release our code.

\section*{Acknowledgments}
\noindent This work was supported by Air Force Office of Scientific Research contract FA9550-17-1-0017.

\bibliography{main}

\newpage
\section{Supplementary Material}
\beginsupplement
\subsection{Algorithms}
\label{sec:algorithms}
Algorithm 1 describes the \textit{Agg-Var} clustering algorithm used as part of CBCL to generate centroid pairs for the RGB-D images. Algorithm 2 explains the category merging process using the Silhouette indices of the training images of all the categories. 

\begin{algorithm}
\caption{\textit{Agg-Var} Clustering}
\begin{flushleft}
        \textbf{Inputs:}
        $F_{RD}$ \textbf{:} RGB and depth feature map pairs from a training dataset with $m$ categories and $N$ samples\\
        $D$\textbf{:} Distance threshold hyperparamter\\
        $w_{rgb}$, $w_D$ \textbf{:} RGB and depth fusion weights hyperparameters\\
        \textbf{Output:} A collection containing a set of centroid pairs for each category of indoor scene, $C_{RD}={C_{RD}^1,C_{RD}^2,...,C_{RD}^m}$
\end{flushleft}
\begin{algorithmic}[1]
\State $F_{RD}^j$\textbf{:} the set of RGB and depth feature map pairs labeled as category $j$, with $N_j$ samples, where $f_{rd_i}^j$ represents the $i$-th sample in $F_{RD}^j$.
\State $C_{RD}^j$\textbf{:} set of RGB and depth centroid pairs for category $j$, where $c_{rd_i}^j$ represents the $i$th centroid pair in $C_{RD}^j$.
\For {$j=1$; $j \leq m$} $C_{RD}^j\leftarrow\{f_{rd_i}^j\}$
\EndFor
\For {$j=1$; $j\leq m$}
\For {$i=2$; $i \leq N_j$}
\State $d_{min} \leftarrow \min_{l=1,..,size(C_{RD}^j)} dist_{RD}(c_{rd_i}^j,f_{rd_i}^j)$
\State $x \leftarrow \argmin_{l=1,..,size(C_{RD}^j)} dist_{RD}(c_{rd_i}^j,f_{rd_i}^j)$
\State \textbf{Set} $c_{rd_x}^{j}$ to be the nearest centroid pair
\State \textbf{Set} $w_{x}^{j}$ to be the number of images clustered 
\State in the $x$th centroid pair of category $j$
\If {$d_{min}<D$}
\State Use Eq. 2 to update centroid pair $c_{rd_x}^{j}$
\Else
\State $C_{RD}^j.append(f_{rd_i}^j)$
\EndIf
\EndFor
\EndFor
\end{algorithmic}
\end{algorithm}

\begin{algorithm}
\caption{Category Merging}
\begin{flushleft}
        \textbf{Input:}
        $C = \{C_{RD}^1,...,C_{RD}^m\}$\Comment{current class centroid pair sets}\\
        $F_{RD}$ \textbf{:} RGB and depth feature map pairs from a training dataset with $m$ categories and $N$ samples\\
        %\textbf{require:} $S = \{s_1^1,...,s_i^j,...,s_N^m\}$\Comment{silhouette values for $N$ RGB-D images for all $m$ categories}
        %\textbf{require:} $Y = \{y_1,...,y_N\}$\Comment{Label of closest centroid corresponding to each image}\\
        \textbf{Output:} $Y_{merged} = \{1,...,t\}$\Comment{New labels of $N$ images belonging $t\leq m$ categories}\\
\end{flushleft}
\begin{algorithmic}[1]
\State Repeat Until $z_{conf}^j<0.25 \lor Y_{conf}^j\neq Y_{conf}^{Y_{conf}^j}$
\State Calculate $S = \{s_1^1,...,s_i^j,...,s_N^m\}$\Comment{silhouette values for $N$ RGB-D images for all $m$ categories}
\State Calculate $Y = \{y_1,...,y_N\}$\Comment{Label of closest centroid corresponding to each image}

\For{$j=1$;$j\leq m$}
\State $z_{conf}^j \leftarrow \frac{\sum_{i=1,y=j,s_i^y\leq 0}^{N} s_i^y}{N_j}$\Comment{percentage of images in class $j$ with Silhouette value <=0}
\State $Y_{conf}^j \leftarrow \max_{i=1,...,m} \sum_{k=1,s_k^i\leq 0,y_k=i}^N 1 $\Comment{most common category among images with $s\leq 0$}
\If{$z_{conf}^j>0.25 \land Y_{conf}^j=Y_{conf}^{Y_{conf}^j}$}
\State $Y_{conf}^j=j$\Comment{Merge categories by assigning the same label}
\EndIf
\EndFor
\end{algorithmic}
\end{algorithm}

\newpage
\subsection{Hyperparameter Values}
%For finetuning Places365-CNN on depth data, for both of the datasets, we use a fixed learning rate of 0.0001, cross-entropy loss with minibatches of size 56 and optimize with stochastic gradient descent. 0.79, 130, 9
In order to train VGG16 on depth data, for both of the datasets, we train it for 200 epochs with an initial learning rate of 0.01 and divide it by 50 after 60, 120 and 160 epochs. Stochastic gradient descent with momentum of 0.9 was used with minibatches of size 64 and trained with cross-entropy loss with weight decay of 0.0005. The hyperparameters for CBCL (distance threshold $D$, \textit{fusion weight} $w_D$ and number of closest centroids for classification $n$) were tuned using cross-validation on the training set. For the SUN RGB-D dataset, the values of the hyperparameters $D$, $n$, and $w_D$ are tuned to 85, 17, and 0.73 respectively. For NYU Depth V2 dataset, $D$, $n$, and $w_D$ are tuned to 95, 7, and 0.76, respectively. For both the datasets, $w_R$ is set to 1.0. Even though the hyperparameters for the two datasets are a little different, CBCL is robust to changes in the hyperparameter values for a large range (Section 4.3.2 in the paper). 

\newpage
\subsection{Confusion Matrices}
The confusion matrices for SUN RGB-D and NYU Depth-V2 datasets are shown below:

\begin{figure}[H]
\centering
\includegraphics[scale=0.27]{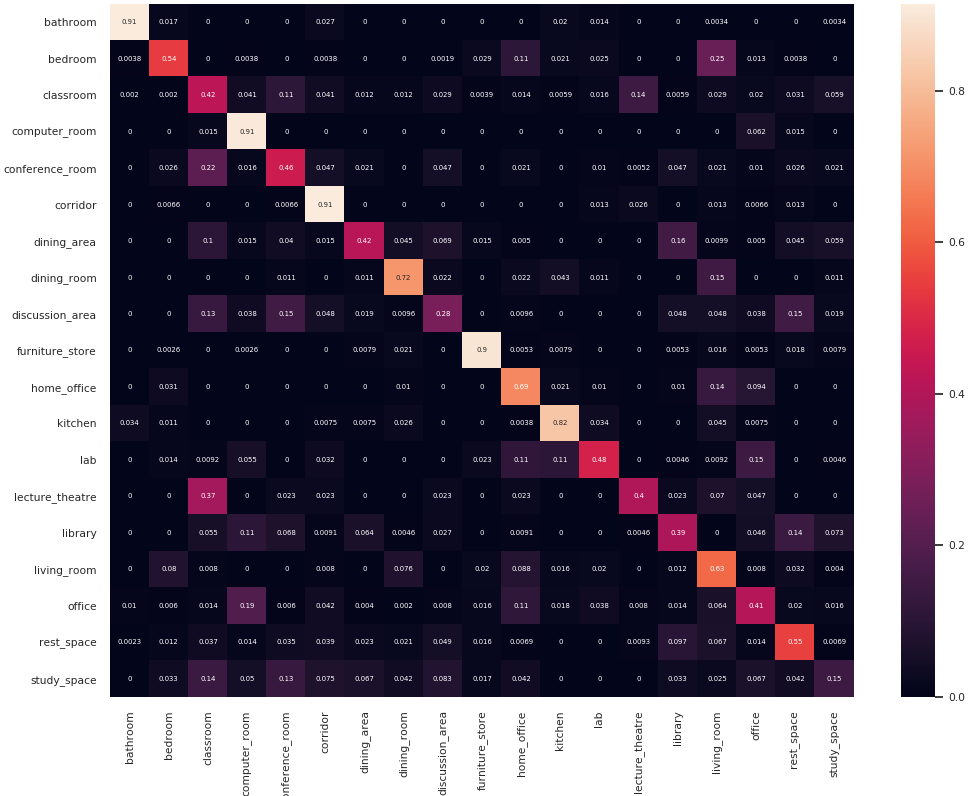}
\caption{\small The confusion matrix for CBCL on the SUNRGB-D dataset. The vertical axis depicts the ground truth and the horizontal axis dpicts the predicted labels.}
\label{fig:confusion_SUN}
\end{figure}

\begin{figure}[H]
\centering
\includegraphics[scale=0.27]{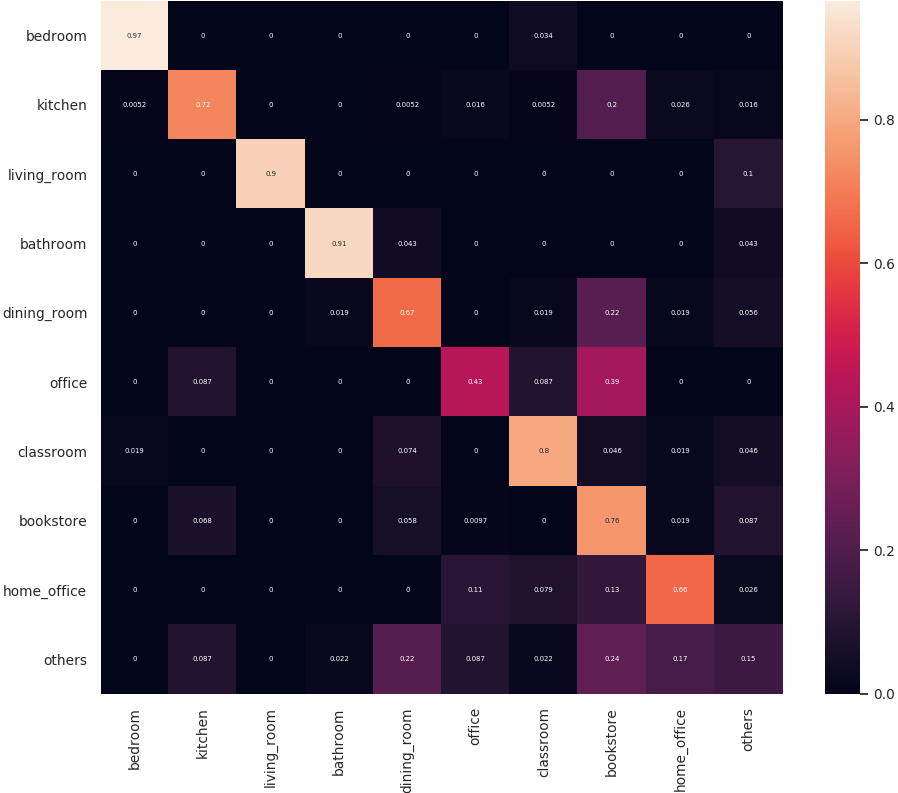}
\caption{\small The confusion matrix for CBCL on the NYU Depth V2 dataset. The vertical axis depicts the ground truth and the horizontal axis depicts the predicted labels.}
\label{fig:confusion_NYU}
\end{figure}

\subsection{Ablation Study on NYU Depth-V2 Dataset}
Table \ref{tab:NYU_ablation} presents the results for an ablation study on the NYU Depth-V2 dataset. We compare CBCL against the same three baselines (VGG, Alexnet and ResNet18) and two clustering hybrids (Agglomerative and K-means) as for SUN RGB-D dataset. Similar to the ablation study for SUN RGB-D dataset, the results indicate that most of the performance gain for our approach comes from better classification of RGB features and \textit{Agg-Var} clustering. 

\begin{table}[H]
\centering
\small
\begin{tabular}{ |p{3.9cm}|p{1.0cm}|p{1.0cm}|p{1.0cm}| }
     \hline
    \textbf{Methods} & \textbf{RGB} & \textbf{Depth} & \textbf{Fusion} \\
    \hline
    VGG Baseline & 57.34 & 49.15 & 60.18\\
    \hline
    Alexnet Baseline & 59.50 & 49.30 & 60.60\\
    \hline
    ResNet18 Baseline & 59.80 & 52.30 & 63.80\\
    \hline
    Agglomerative RGB-D & 63.27 & 45.46 & 66.30\\
    \hline
    K-means RGB-D & 61.13 & 43.02 & 63.97\\
    \hline
    \textbf{CBCL} & \textbf{66.40} & 49.50 & \textbf{70.91}\\%\textbf{69.70}\\
 \hline
 \end{tabular}
 \bigskip
 \caption{Ablation study results on the NYU Depth-V2 dataset}
 \label{tab:NYU_ablation}
 \end{table}

\subsection{Examples of Conceptually Similar Categories}

\begin{figure}[h]
\centering
\includegraphics[scale=0.285]{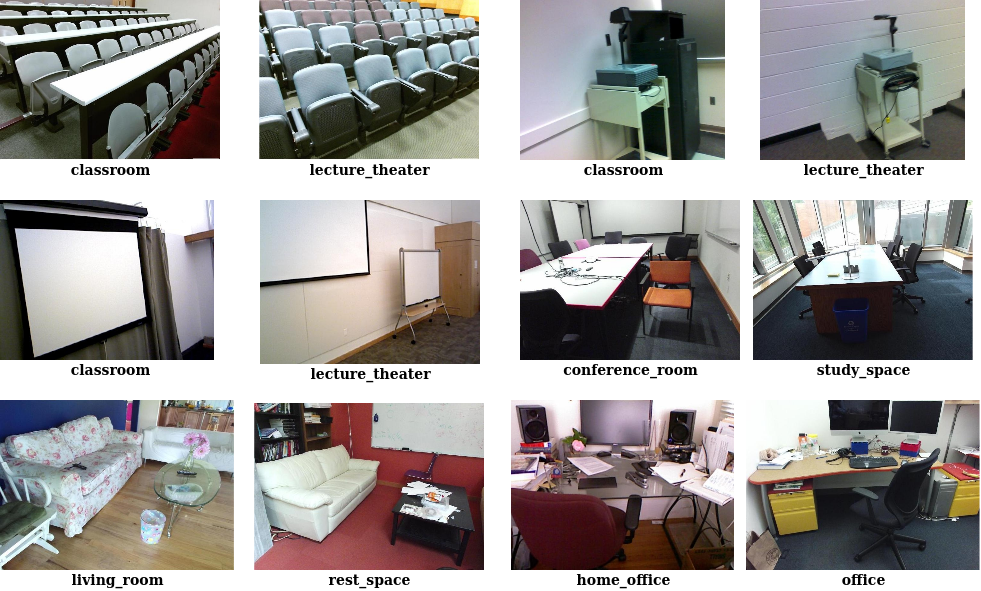}
\caption{\small Additional examples of images from conceptually similar scene categories. These images are from the SUN RGB-D dataset.}
\label{fig:diff_cetroids_more}
\end{figure}

\subsection{Silhouette Index Values and Category Merging on the SUN RGB-D Dataset}
\label{SUN_Category_Merging}
Table \ref{tab:sil_sun} presents the result from calculating the silhouette index values for each image in each category of the SUN RGB-D dataset. The table lists the percentage of images with silhouette value less than or equal to zero and the category most often confused. The italicize categories have greater than 25\% of images with $s<=0$ and were confused bi-directionally, i.e. images with $s\leq 0$ of one category of the pair were mostly confused with the other category in the pair and vice versa. The algorithm for merging the categories is described in Section \ref{sec:algorithms}. The procedure is repeated recursively and stopped if any of the two criteria are not met: categories should have more than 25\% images with $s\leq 0$ and they should be confused bi-directionally.

%a large percentage of images with $s<=0$ are candidates for being merged with other categories.  

%indicate that they could be merged with other categories. However, only those category pairs are combined which have near 30\% images with $s<=0$ and most of these images are confused with the other category in the pair. Only four category pairs satisfy these two conditions: (classroom,lecture\_theatre) , (conference\_room,study\_space), (home\_office,office) and (living\_room,rest\_space). Category pair (dining\_room,discussion\_area) satisfy the second criteria (most of their images with $s<=0$ are confused with other category of the pair) but the total percentage of images of these categories with $s<=0$ are $\sim$13\%. Another category library has a large amount of images (22.3\%) with $s<=0$ but these images are mostly confused with category rest\_space, while rest\_space is most confused with living\_room. Hence, we do not combine these two categories. 

\begin{table}[H]
\centering
\small
\begin{tabular}{ |p{2.1cm}|p{4cm}|p{3.1cm}| }
     \hline
    \textbf{Categories} & \textbf{Percent of images with $s$<=0} & \textbf{Confused Category} \\
    \hline
    bathroom & 4.0  & bedroom\\
    \hline
    bedroom & 8.1  & living\_room\\
    \hline
    \textit{classroom} & \textit{28.5}  & \textit{lecture\_theatre}\\
    \hline
    computer\_room & 7.6 & classroom\\
    \hline
    \textit{conference\_room} & \textit{27.9} & \textit{study\_space}\\
    \hline
    corridor & 11.7 & bedroom\\
    \hline
    dining\_area & 12.8 & rest\_space\\
    \hline
    dining\_room & 13.8 & discussion\_area\\
    \hline
    discussion\_area & 12.4 & dining\_room\\
    \hline
    furniture\_store & 7.4 & bedroom\\
    \hline
    \textit{home\_office} & \textit{29.2} & \textit{office}\\
    \hline
    kitchen & 6.8 & furniture\_store\\
    \hline
    lab & 9.5 & kitchen\\
    \hline
    \textit{lecture\_theatre} & \textit{32.0} & \textit{classroom}\\
    \hline
    library & 22.3 & rest\_space\\
    \hline
    \textit{living\_room} & \textit{30.4} & \textit{rest\_space}\\
    \hline
    \textit{office} & \textit{28.4} & \textit{home\_office}\\
    \hline
    \textit{rest\_space} & \textit{29.7} & \textit{living\_room}\\
    \hline
    \textit{study\_space} & \textit{31.9} & \textit{conference\_room}\\
 \hline
 \end{tabular}
 \bigskip
 \caption{Percentage of images in each of the 19 categories of SUN RGB-D dataset that have silhouette's index $s<=0$ and the corresponding categories with which most of these low silhouette value images are closer to.}
 \label{tab:sil_sun}
 \end{table}

\subsection{Category Merging for NYU Depth V2 Dataset}
The NYU Depth V2 dataset includes a large number of categories with a few training images which traditionally are combined into one "others" category \cite{Silberman12} to reduce the total number of categories from 27 to 10. The 17 categories that typically compose the "others" category, however, are conceptually different from each other. The silhouette index values were calculated on all 27 categories. The same criteria was used to identify mergeable categories for the SUN RGB-D dataset. The recursive analysis indicated that the following category should be merged: (home\_office, foyer, study, basement), (dining\_room, dinette), (cafe, bookstore, furniture\_store), (kitchen, office\_kitchen), (study\_room, office, printer\_room), (computer\_lab, conference\_room, classroom), (living\_room, playroom, student\_lounge, reception\_room). The categories exercise\_room, home\_storage, indoor\_balcony and laundry\_room did not meet the criteria for merging so they remained in the "others" category. The result was a total of 10 categories but under a different merging scheme from that used by \cite{Silberman12}.    

On this updated dataset, CBCL achieves \textbf{78.95\%} mean class accuracy which is 8.04\% higher than on the original dataset. The VGG baseline achieves 67.24\% accuracy which is 7.06\% higher than on the original dataset, although still 11.71\% lower than CBCL. These results are similar to the SUN RGB-D results.

\end{document}